\documentclass[letterpaper, 10 pt, conference]{ieeeconf} 
\IEEEoverridecommandlockouts                              %
\usepackage[pdftex]{graphicx}
\usepackage{url}
\usepackage{hyperref}

\usepackage[usenames, dvipsnames]{color}
\usepackage{soul}

\usepackage{mathptmx} %
\usepackage{times} %
\usepackage{amsmath} %
\usepackage{amssymb}  %
\usepackage[noadjust]{cite}
\usepackage{bm}
\usepackage[per-mode=symbol]{siunitx}
\usepackage{pgfplots} 
\usepackage{pgfgantt}
\usepackage{pdflscape}
\usepackage{outlines}

\usepackage{fancyhdr}
\newcommand{\mytitle}{\textbf{Accepted final version.}
To appear in \textit{Proceedings of the IEEE}.\\
\copyright 2022 IEEE. Personal use of this material is permitted. Permission
from IEEE must be obtained for all other uses, in any current or future
media, including reprinting/republishing this material for advertising or
promotional purposes, creating new collective works, for resale or
redistribution to servers or lists, or reuse of any copyrighted component of
this work in other works.}

\title{\LARGE \bf
Fast Reflexive Grasping with a Proprioceptive Teleoperation Platform
}

\author{Andrew SaLoutos$^{1}$, Elijah Stanger-Jones$^{1}$, and Sangbae Kim$^{1}$ %
\thanks{$^{1}$Authors are with the Biomimetic Robotics Laboratory at the Department of Mechanical Engineering, Massachusetts Institute of Technology (MIT), Cambridge, MA, 02139, USA. {\tt\small\url{saloutos@mit.edu}}} }

\begin{document}
\bstctlcite{IEEEexample:BSTcontrol}

\maketitle
\thispagestyle{fancy}
\fancyhf{}		%
\fancyfoot[L]{\normalfont \sffamily  \scriptsize \mytitle}		%
\addtolength{\footskip}{-10pt}    %
\pagestyle{empty}

\begin{abstract}

We present a proprioceptive teleoperation system that uses a reflexive grasping algorithm to enhance the speed and robustness of pick-and-place tasks.
The system consists of two manipulators that use quasi-direct-drive actuation to provide highly transparent force feedback. 
The end-effector has bimodal force sensors that measure 3-axis force information and 2-dimensional contact location.
This information is used for anti-slip and re-grasping reflexes.
When the user makes contact with the desired object, the re-grasping reflex aligns the gripper fingers with antipodal points on the object to maximize the grasp stability. 
The reflex takes only 150ms to correct for inaccurate grasps chosen by the user, so the user's motion is only minimally disturbed by the execution of the re-grasp. 
Once antipodal contact is established, the anti-slip reflex ensures that the gripper applies enough normal force to prevent the object from slipping out of the grasp.
The combination of proprioceptive manipulators and reflexive grasping allows the user to complete teleoperated tasks with precision at high speed. 

\end{abstract}

\section{Introduction}

In teleoperation tasks, matching the device and controller design to the capabilities of the user can significantly improve performance. Most traditional robot arms, however, have sluggish dynamics compared to a human's control bandwidth.
To enable fast and accurate teleoperation, the robotic system needs to be transparent: the force and velocity of the end-effectors of the input (leader) and output (follower) devices should be identical. 
With no communication delay and perfect sensing, the transparency can be accomplished by directly measuring the forces and torques at the end-effector of the follower device and applying them to the user, while the follower device perfectly tracks the position and velocity of the leader device. 
However, in practice, time delays and noisy sensing can cause instabilities, leading to slower and more conservative teleoperation \cite{niemeyer2016telerobotics}. 

By employing the proprioceptive actuation paradigm, the closed-loop control bandwidth of the teleoperation system can be increased and good transparency can still be achieved. Direct-drive arms were first introduced by Asada \cite{asada1983}, and the design principles have been successfully applied to other manipulators, such as the WAM arm, the Phantom haptic interface, and the Ambidex arm \cite{townsend1993mechanical, massie1994phantom, kim2017anthropomorphic}. These manipulators have low link inertia, low reflected inertia, and relatively high torque density, enabling fast and accurate open-loop force control. Proprioceptive grippers have also been developed by Bhatia \cite{bhatia2019direct} and Lin \cite{lin2021exploratory}, since transparency at the fingers and hand is just as important as transparency throughout the arm for everyday tasks. However, extending these designs to match the full dexterity of a human wrist and hand will be exceedingly difficult.

To preserve the benefits of proprioceptive actuation without adding too much complexity, autonomous reflexes can be added to the follower system to assist the teleoperator during grasping. In this case, the user still feels a transparent system, but they do not have to provide highly accurate position and force commands for each grasp. For pick-and-place teleoperation, two potential ways to assist the user are to prevent object slip during grasps and to autonomously re-grasp objects to ensure stable grasp configurations. 

\begin{figure}[t]
\centering
\includegraphics[width=0.9\linewidth]{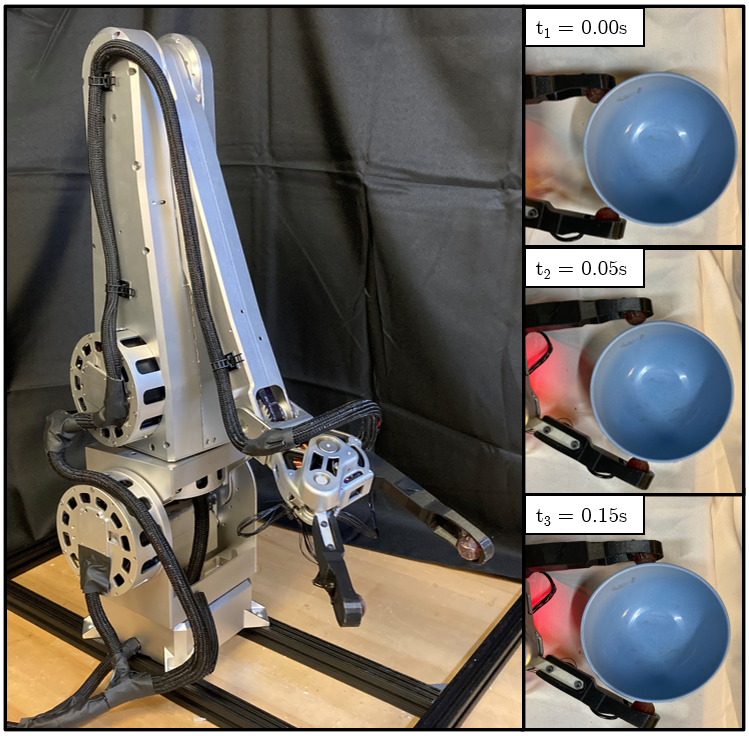}
\caption{ \textbf{Manipulator and re-grasp attempt.} High torque density motor modules and low-inertia links make the new arm a good platform for high-bandwidth teleoperation. Bimodal force sensors on the gripper enable reflexive grasping. The inset frames 1 through 3 show a single re-grasp attempt, which takes just 150ms and achieves a secure antipodal grasp. }
\label{fig:arm_with_timelapse}
\vspace{-4mm}
\end{figure}

An extensive overview of previous work on slip detection is given in a survey by Romeo \cite{romeo2020methods}. Many previous methods have used direct force measurements \cite{melchiorri2000slip, bicchi1989augmentation}, which can result in instability when used in closed-loop control. Gunji et al developed a feedback controller for slip prevention that was constructed around analog pressure measurements, but the controller had a response time of over 0.5s \cite{gunji2008grasping}. Other recent works use learning methods to measure forces or learn slip detection directly \cite{huh2020dynamically, li2018slip}. Overall, these algorithms are limited by low sensor bandwidth and are not able to react to slip caused by sudden impacts or fast motions. 

Re-grasping has also been a long-standing area of interest in robotic manipulation \cite{tournassoud1987regrasping}, referring to changing in-hand grasp poses or forces \cite{schlegl2001fast} or trying again after unsuccessful grasps \cite{calandra2018more}. In several works, learning methods have been applied to evaluate grasp quality metrics, which are then used to decide if a re-grasp attempt is necessary \cite{calandra2018more, murali2018learning, chebotar2016self}. However, the response time of many of these re-grasping algorithms is slow compared to human manipulation and would hinder teleoperation performance. 

\textbf{Contributions.} This work contains two main contributions. First, we introduce a proprioceptive teleoperation system consisting of two human-scale arms specifically designed for high-bandwidth manipulation. Second, we develop two autonomous reflexes that are combined to improve users' grasping accuracy and speed. We also present experiments showing that the combination of a proprioceptive manipulator and fast reflexive grasping controllers results in fewer failed grasp attempts and faster average trial times for teleoperated pick-and-place tasks.

\section{Teleoperation Platform} \label{sec:teleop}

\begin{figure}[t]
\centering
\includegraphics[width=0.9\linewidth]{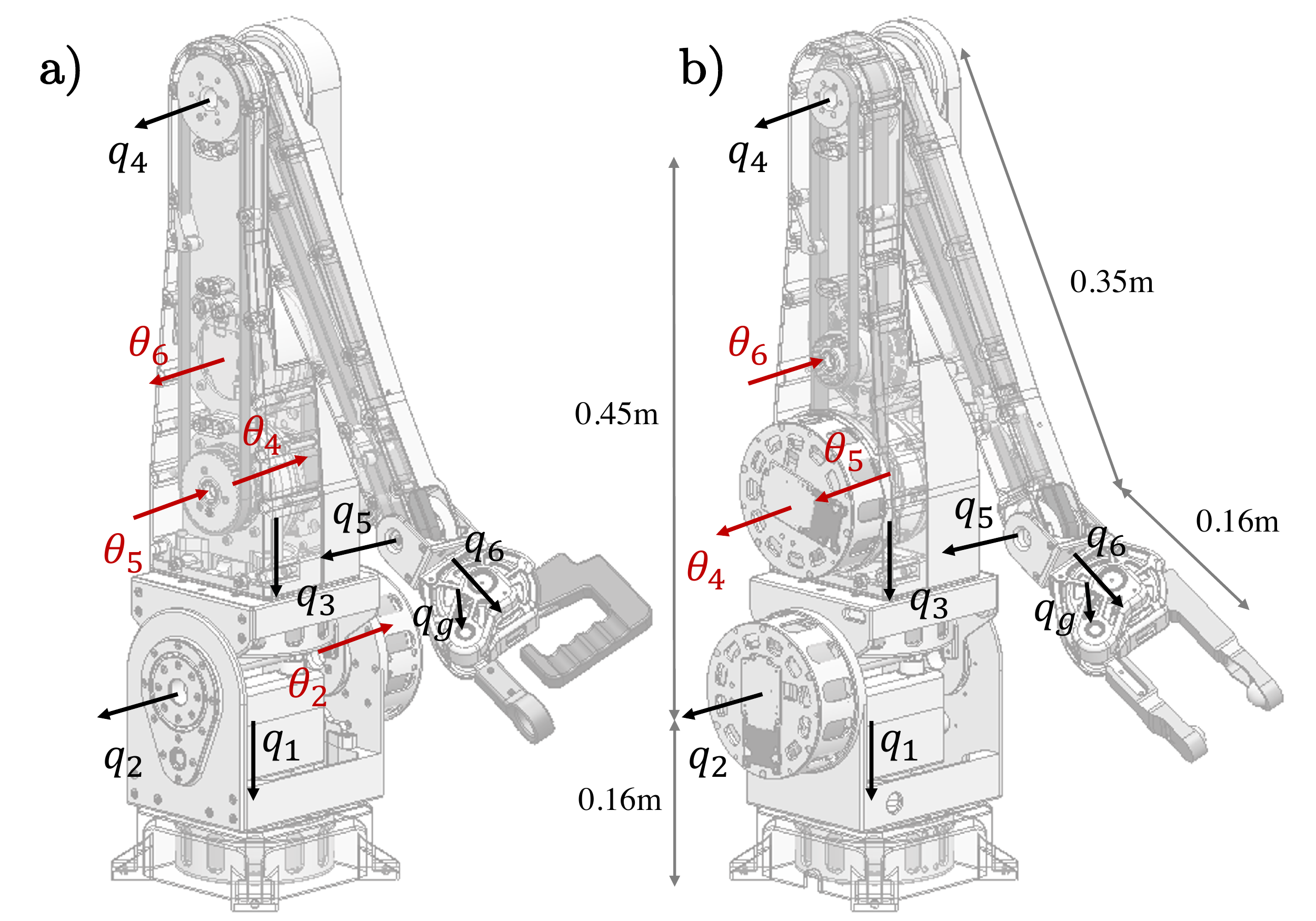}
\caption{\textbf{Teleoperation platform}. (a) Leader arm with input gripper. (b) Follower arm with sensorized gripper. Joint axes are represented by $q_i$. Actuator axes are represented by $\theta_i$ if the actuator is not located directly at the joint. Each vector corresponds to the axis of positive rotation.}
\label{fig:arm_axes}
\vspace{-2mm}
\end{figure}

One of the quasi-direct-drive manipulators is shown in Fig. \ref{fig:arm_with_timelapse}.
The teleoperation platform consists of two of these manipulators, with specialized grippers for the leader and follower devices, which are shown in Fig. \ref{fig:arm_axes}.
Each arm has 6 degrees of freedom, a total reach of 0.96m, and a maximum payload of 5kg while having a total mass of just 13.8kg.
To minimize the effective mass at the wrist and thus increase the control bandwidth, the arm's actuators are all placed as close as possible to the base.
Both arms are controlled simultaneously over SPI at 500Hz by an Intel-i7 single board computer. A custom control PCB converts the SPI commands into CAN commands for each actuator which are sent at up to 3kHz.

The high torque density actuators provide equivalent payload capabilities compared to other common commercial robotic manipulators with a lower overall mass \cite{townsend1993mechanical,kuka,ur5}.
The high speed capability and low inertia of the arms increase the responsiveness of the teleoperation system to user inputs, at the cost of worse position repeatability. 

\subsection{Modular Actuators}

The manipulator is designed around two types of high torque density brushless DC motor modules, based on the design of the actuators used in the MIT Mini Cheetah \cite{katz_mini}, which are summarized in Table \ref{tab:actuator_comp}. The larger actuator is based on the U12-II motor, and the smaller actuator is based on the U10 motor, both from the T-motor company. 
The actuators have integrated motor drivers that are a modified version of the ones used in the Mini Cheetah.

\begin{table}[t]
\caption{Comparing Modular Actuators}
\label{tab:actuator_comp}
\centering
\begin{tabular}{ c  c  c } 
 \hline
Module & U12 & U10 \\
 \hline\hline
Mass (\si{g}) & 1170 & 620 \\ 
 \hline
Gear Ratio & 6:1 & 6:1  \\
 \hline
Outer Diameter (\si{mm}) & 120 & 101 \\
 \hline
Thickness (\si{mm}) & 44 & 36 \\
 \hline
Peak Torque (\si{Nm}) & 72.0 & 34.0 \\
 \hline
Max. Speed at 60\si{V} (\si{rad/s}) & 39.0 & 48.0 \\
 \hline
Rotor Inertia (\si{kgm}$^2$) & 5e-5 & 3e-5 \\
 \hline
 \\
\end{tabular}
\vspace{-6mm}
\end{table}

\subsection{Arm Design}
The degrees of freedom in each arm are labeled in Fig. \ref{fig:arm_axes}.
The shoulder and elbow joints, $q_1$ through $q_4$, use the U12 actuators and the wrist joints, $q_5$ and $q_6$, use the U10 actuators. 

The mass of the shoulder joint is 6.55kg, or roughly half the mass of the total arm. To reduce the torque requirement of the shoulder pitch motor and improve the payload capacity of the arm, a compression spring counterbalance is added in the center of the shoulder joint. The counterbalance compensates for the mass of the shoulder roll joint and the entire upper arm while keeping the shoulder joint as compact as possible. Across the range of motion of the shoulder pitch joint, the counterbalance compensates for at least $68\%$ of the upper arm mass.

The upper arm link contains the actuators for the elbow, wrist pitch, and wrist roll axes. 
High-stiffness timing belt transmissions are used for each of the joints. The elbow transmission uses a single belt and each wrist joint uses two belts, with floating shafts on either side of the elbow joint. The belt transmissions do not provide any additional reductions, but there is a 2:1 bevel gear stage for the wrist roll joint axis. The mass of the upper arm link is 4.95kg, which accounts for 38\% of the total arm mass. The mass of the lower arm link and wrist joint assembly is 1.1kg, which is just 8\% of the total mass. 

The gripper is shown in detail in Fig. \ref{fig:gripper_sensors}(a). It has one fixed finger and one actuated finger, $q_g$ in Fig. \ref{fig:arm_axes}, and is built around a frameless version of the R60 motor from T-motor. There is a 2:1 belt reduction between the motor and the finger to bring the maximum grasp force to approximately 50N, which corresponds to lifting the maximum payload of the arm. The gripper is approximately 200mm in length, can grasp objects up to 95mm in diameter, and has a total mass of 590g.

\subsection{Bimodal Fingertip Force Sensors}

At the tip of each gripper finger is a bimodal force sensor consisting of eight BMP388 barometric pressure sensors under a dome of polyurethane \cite{epstein2020}. The sensors saturate at approximately 25N of normal force and have a force resolution of less than 0.5N. Each sensor is 20mm in diameter and has a maximum sampling rate of 200Hz. 

\begin{figure}[t]
\centering
\includegraphics[width=\linewidth]{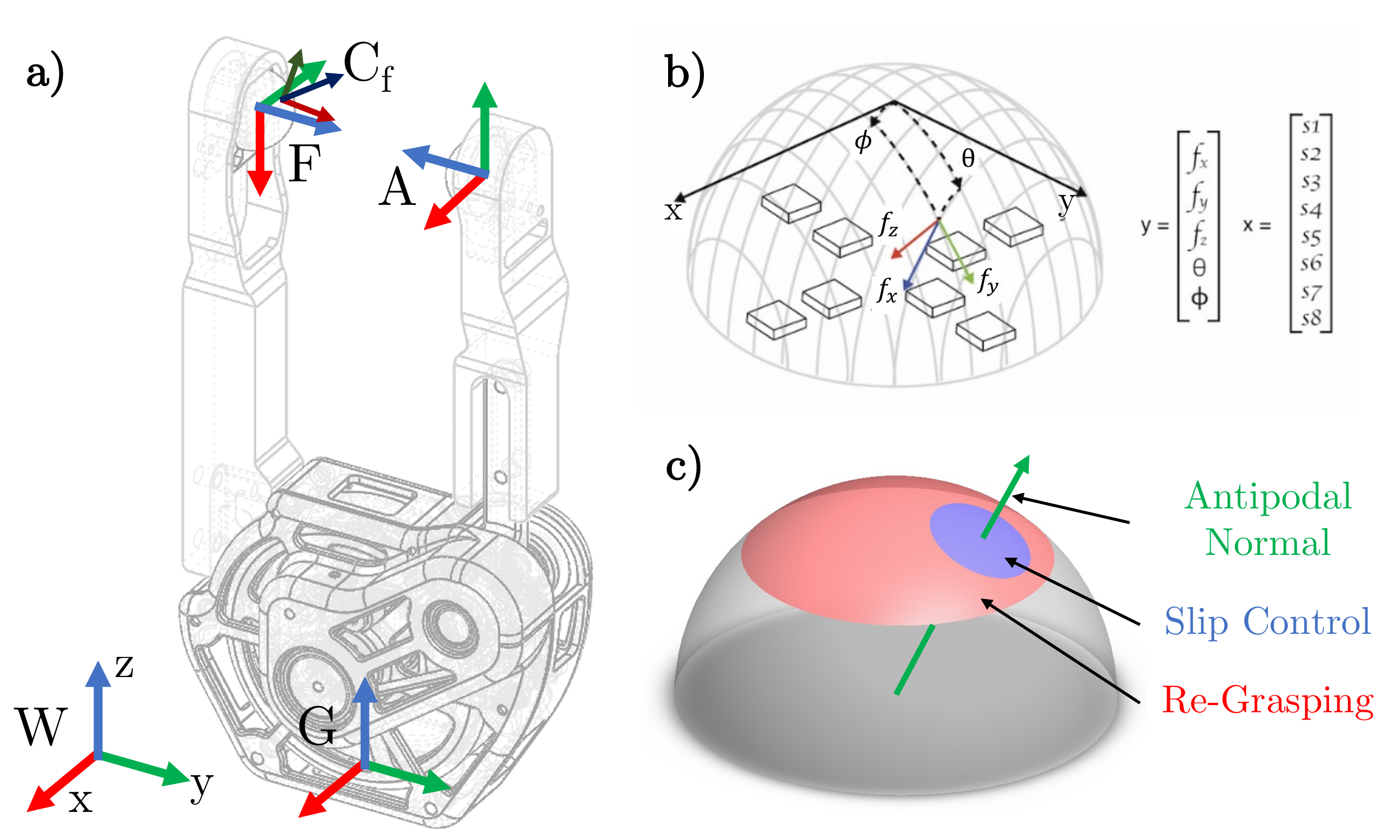}
\caption{\textbf{Sensorized gripper.} (a) Definitions of gripper frames for reflex calculations. (b) Parameterization of the sensor surface, along with the neural network inputs, $x$, and outputs, $y$. (c) Zones of the sensor surface correspond to different algorithm states. Re-grasping can occur from anywhere on the sensor's usable surface, while the anti-slip reflex is only enabled in the region near the antipodal normal vector.}
\label{fig:gripper_sensors}
\vspace{-4mm}
\end{figure}

For each sensor, a neural network is trained to provide estimates of the contact location and contact forces based on the readings of the individual pressure sensors. Fig. \ref{fig:gripper_sensors} shows the sensors at the fingertips (a) as well as a diagram of the pressure sensors and contact coordinates (b). The contact frame, given by (\ref{eq:contact_transform}), is parameterized by two angles, $\theta$ and $\phi$. At the contact location, 3-axis force data is reported as the normal force (along the z-axis) and the tangential or shear forces along the x- and y-axes of the local frame. Due to the parameterization of the contact location, the z-axis of the contact force frame is also the contact normal vector. 

\begin{equation} \label{eq:contact_transform}
\begin{split}
\boldsymbol{T}_{base,contact} &= 
\begin{bmatrix}
\boldsymbol{R}_y(\phi) & \boldsymbol{0} \\
\boldsymbol{0}^T & 1 \\
\end{bmatrix}
\begin{bmatrix}
\boldsymbol{R}_x(\theta) & \boldsymbol{0} \\
\boldsymbol{0}^T & 1 \\
\end{bmatrix}
\begin{bmatrix}
\boldsymbol{I}_3 & \boldsymbol{p}_0 \\
\boldsymbol{0}^T & 1 \\
\end{bmatrix}
\\
\boldsymbol{p}_0 &= \begin{bmatrix} 0 & 0 & r_{sensor} \end{bmatrix}^T
\end{split}
\end{equation}

\section{Reflexive Grasping Algorithm} \label{sec:alg}

\begin{figure}[t]
\centering
\includegraphics[width=0.9\linewidth]{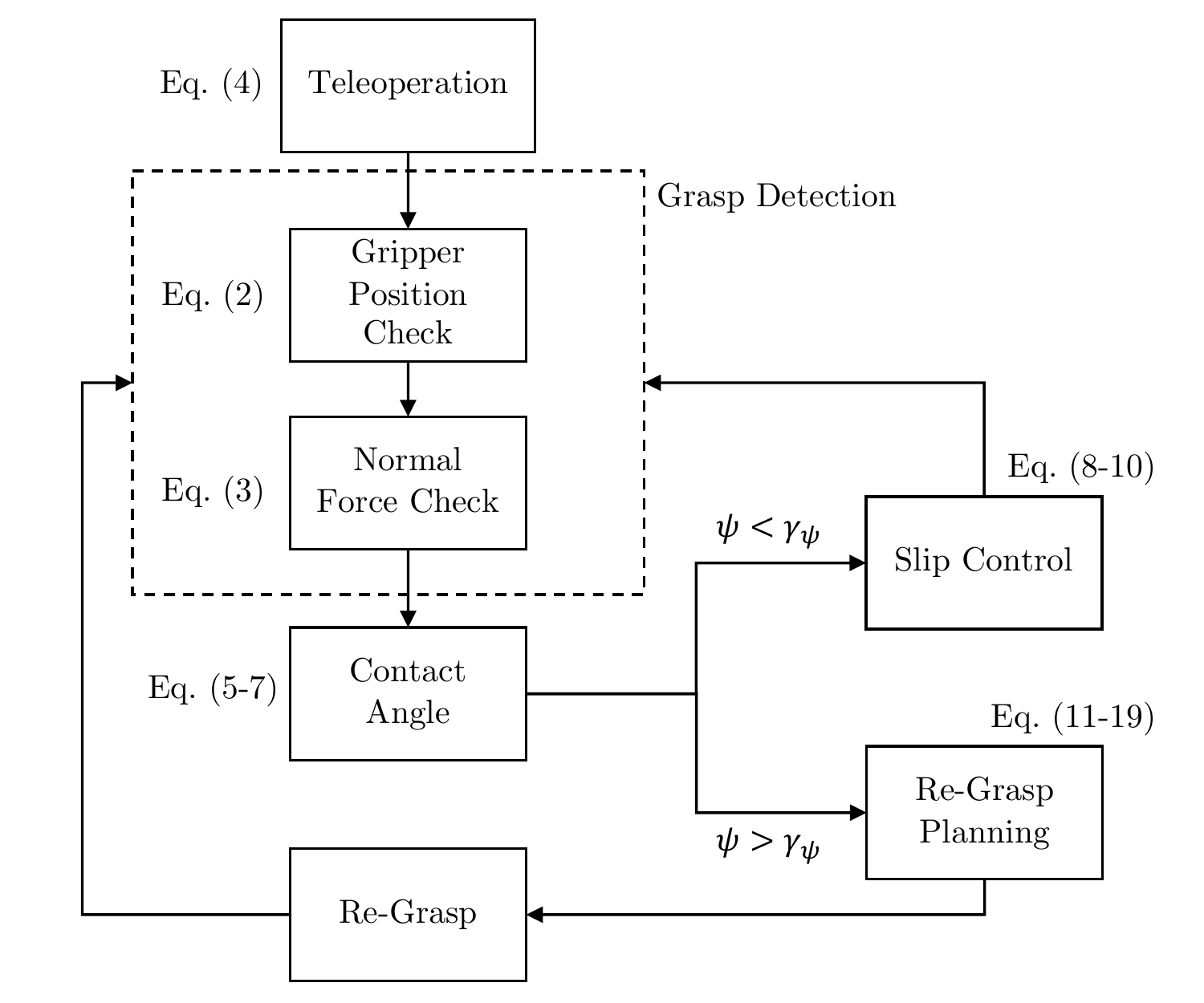}
\caption{\textbf{Reflexive grasping algorithm.} The system starts in a bilateral teleoperation mode, checking the grasp detection conditions at each time-step. If a grasp is detected, the system chooses to activate the anti-slip reflex or the re-grasping reflex.}
\label{fig:alg_flowchart}
\vspace{-4mm}
\end{figure}

A flowchart for our reflexive grasping algorithm is shown in Fig. \ref{fig:alg_flowchart}. On system startup, the arms enter a bilateral teleoperation mode, described in Section \ref{sec:alg:teleop}. Then, at each control time-step, two hierarchical grasp detection conditions are checked. The higher-level condition is based on the leader and follower gripper positions, $q_{g,l}$ and $q_{g,f}$, and is given by

\begin{equation} 
\delta_{gripper} = 
\begin{cases} 
      1 & \text{if } q_{g,l} - q_{g,f} \leq \gamma_q \\
      0 & \text{otherwise} \\
   \end{cases},
\label{eq:gripper_check}
\end{equation}

\noindent where $\gamma_q$ is an adjustable threshold. The second, lower-level condition depends on the normal forces measured by both sensors on the follower gripper, given by

\begin{equation} 
\delta_{F_n} = 
\begin{cases} 
      1 & \text{if } |F_{n,f}| \geq \gamma_{n}, |F_{n,a}| \geq \gamma_{n} \\
      0 & \text{otherwise} \\
   \end{cases},
\label{eq:fn_check}
\end{equation}

\noindent where the subscripts $f$ and $a$ refer to the fixed and actuated gripper fingers, respectively, and $\gamma_{n}$ is another adjustable threshold. If both trigger conditions are met, the system registers the detected grasp. Then, the system checks the contact angles to determine whether to activate the anti-slip reflex or to plan and execute a re-grasping trajectory. The contact angle calculations are described in Section \ref{sec:alg:cont_check}, the anti-slip reflex is described in Section \ref{sec:alg:slip}, and the re-grasping reflex is described in Section \ref{sec:alg:regrasp}. 

One key assumption in the calculations for both reflexes is that the grasped objects can be modeled reasonably well with a circular cross-section; this assumption holds well for many common objects.

\subsection{Teleoperation Architecture} \label{sec:alg:teleop}

The leader and follower arms are mirrored from one another, with positive joint axes shown in Fig. \ref{fig:arm_axes}. In the nominal teleoperation mode, the PD controller in (\ref{eq:teleop_coupling}) is applied between the corresponding joints of each arm, including the grippers, along with a term for gravity compensation. Due to the system's low inertia and fast control bandwidth, this PD coupling can give high-stiffness force feedback to the user.

\begin{equation} 
\begin{split}
\tau_{f,i} &= K_p(q_{l,i} - q_{f,i}) + K_d(\dot{q}_{l,i}-\dot{q}_{f,i}) + \tau_{g}
\\
\tau_{l,i} &= -\tau_{f,i}
\end{split}
\label{eq:teleop_coupling}
\end{equation}

\subsection{Calculating Contact Frames} \label{sec:alg:cont_check}

Once a contact is detected, the SE(3) contact frames for the force sensors are calculated as $\boldsymbol{T}_{F,C_f}$ and $\boldsymbol{T}_{A,C_a}$, where the subscripts $F$ and $A$ correspond to the frames at each fingertip and $C_i$ denotes the calculated contact frame for each sensor. These frames are shown in Fig. \ref{fig:gripper_sensors}(a), along with the gripper base frame, $G$, and the world reference frame, $W$. 

To activate the anti-slip reflex, the contacts need to be nearly antipodal, or the object will be pushed out of the gripper. To check this condition, two antipodal desired contact frames, $\boldsymbol{T}_{F,C_f^*}$ and $\boldsymbol{T}_{A,C_a^*}$, are constructed using the gripper kinematics. Fig. \ref{fig:slip_diagrams}(a) shows an example of measured contacts and the calculated antipodal contact thresholds for the current gripper configuration. 

Once the desired contact frames are constructed, the measured contact frames can be converted into the desired contact frames according to (\ref{eq:c_to_cdes}). With these transforms, the polar angle $\psi$ between the z-axis of the desired contact frame and the measured contact normal can be calculated according to (\ref{eq:psi_calc}).

\begin{equation} 
\begin{split}
\boldsymbol{T}_{C_f^*,C_f} = \boldsymbol{T}_{F,C_f^*}^{-1}\boldsymbol{T}_{F,C_f} \\
\boldsymbol{T}_{C_a^*,C_a} = \boldsymbol{T}_{A,C_a^*}^{-1}\boldsymbol{T}_{A,C_a}
\end{split}
\label{eq:c_to_cdes}
\end{equation}

\begin{equation} 
\psi_i = \arccos(\boldsymbol{T}_{C_i^*,C_i}[2,2]) \text{ for } i \in \{f,a\}
\label{eq:psi_calc}
\end{equation}

If both measured polar angles satisfy the condition

\begin{equation} 
\delta_{\psi} = 
\begin{cases} 
      1 & \text{if } \psi_f \leq \gamma_{\psi}, \psi_a \leq \gamma_{\psi} \\
      0 & \text{otherwise} \\
   \end{cases},
\label{eq:polar_check}
\end{equation}

\noindent where $\gamma_\psi$ is a positive threshold, the contacts are close enough to antipodal to enter the slip control state. If the condition is not met, a re-grasp must be attempted. 

\subsection{Anti-Slip Reflex} \label{sec:alg:slip}

\begin{figure}[t]
\centering
\includegraphics[width=\linewidth]{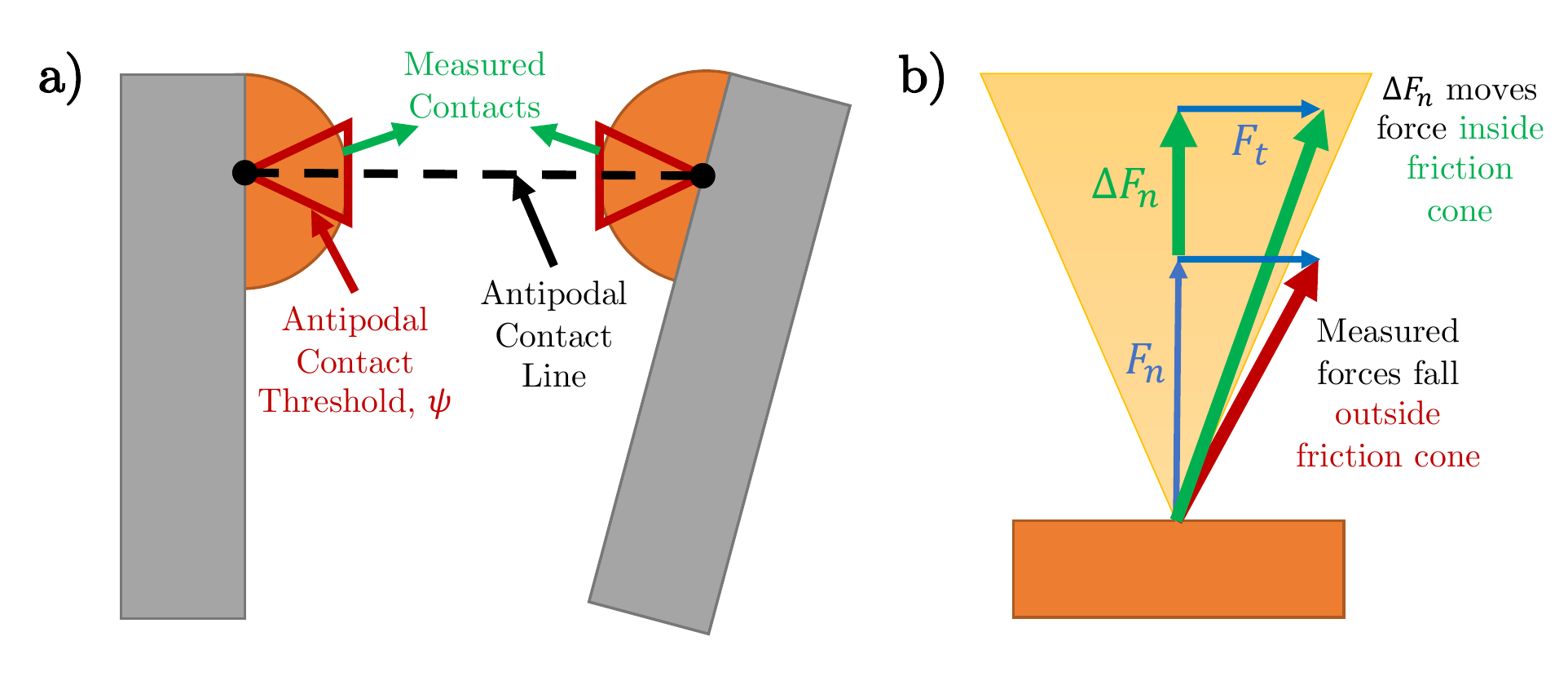}
\caption{\textbf{Contact checking and anti-slip reflex.} (a) The antipodal threshold is used to determine whether to activate the anti-slip reflex. (b) Anti-slip principle. Given measured normal and tangential forces, $\Delta F_n$ can be calculated to keep the resultant force within the friction cone.}
\label{fig:slip_diagrams}
\vspace{-4mm}
\end{figure}

The key principle of the anti-slip reflex is shown in Fig. \ref{fig:slip_diagrams}(b). If both polar angles are within the antipodal threshold, and enough force is applied, the contacted object can be stably grasped. Given a measured normal force $F_n$ and a measured shear force $F_t = \sqrt{F_x^2 + F_y^2}$, the desired magnitude of normal force to prevent slip can be calculated as

\begin{equation} 
|F_{n,des}| = \frac{F_t}{\gamma_{\mu}}
\label{eq:slip_control}
\end{equation}

\noindent where $\gamma_{\mu}$ is an adjustable coefficient of friction. This parameter can be decomposed as 

\begin{equation} 
\gamma_{\mu} = \frac{\hat{\mu}}{\gamma_{c}}
\label{eq:fric_cone_thresh}
\end{equation}

\noindent where $\hat{\mu}$ is an estimate of the coefficient of friction and $\gamma_{c} \geq 1$ controls how close the resulting force will stay to the edge of the estimated friction cone. If this estimate is selected conservatively the system will grip objects harder than necessary to prevent slip, but the overall performance will not be affected. If this parameter is set near or above the real coefficient of friction, grasps will begin to fail.

A desired normal force is calculated based on measurements from each sensor, and then the larger of the two forces is applied as a feed-forward gripper torque as in (\ref{eq:tauff}). Since the gripper has low finger inertia and is nearly direct-drive, additional feedback control is not necessary to achieve the desired normal force.

\begin{equation} 
\tau_{g,s} = -|F_{n,des}|l_{finger}.
\label{eq:tauff}
\end{equation}

\subsection{Re-Grasping Reflex} \label{sec:alg:regrasp}

\begin{figure}[t]
\centering
\includegraphics[width=\linewidth]{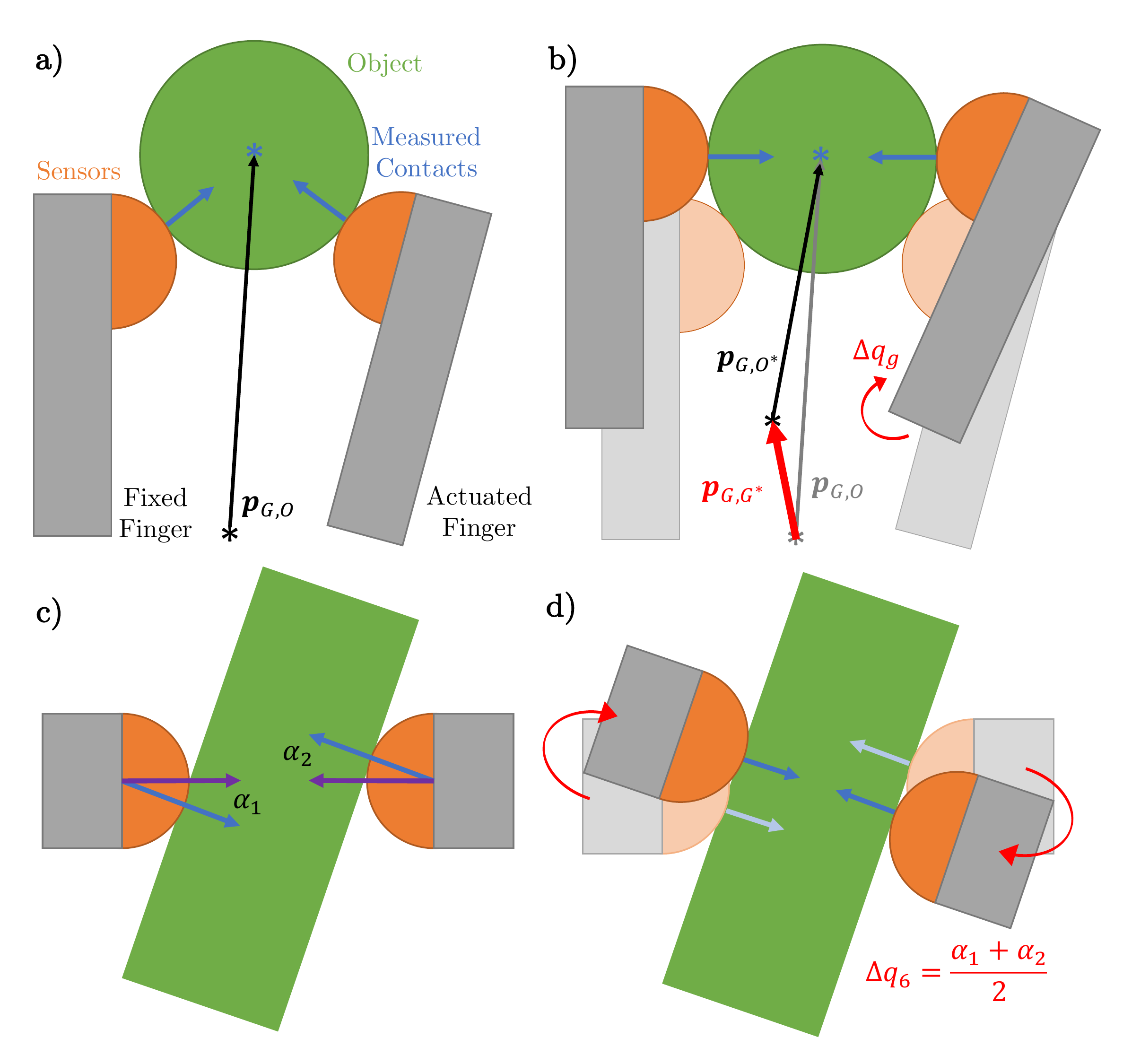}
\caption{\textbf{Stages of re-grasping.} During re-grasping, adjustments to the gripper pose and orientation are calculated using the contact kinematics. (a) Calculating the object center location. (b) Calculating the desired gripper base position for antipodal contact. (c) A case of out-of-plane misalignment between the object and the gripper. (d) Calculating a wrist rotation that will correct the misalignment.}
\label{fig:regrasp_diagrams}
\vspace{-4mm}
\end{figure}

If the contacts are not antipodal, the re-grasping reflex is activated. First, the contact information from both sensors is used to estimate the location of the object center, as shown in Fig. \ref{fig:regrasp_diagrams}(a). Assuming an object with an approximately circular cross-section, the center of the object, $\boldsymbol{p}_{G,O}$, should be an equal distance (its radius, $r$) along each contact normal, $\boldsymbol{n}_{C}$, from each contact location, $\boldsymbol{p}_{G,C}$. This relation is given by 

\begin{equation}
\boldsymbol{p}_{G,O} = \boldsymbol{p}_{G,C_i} + r\boldsymbol{n}_{C_i} \text{ for } i \in \{f,a\}.
\label{eq:pwo}
\end{equation}

Splitting both of these vector equations into their components yields 6 linear equations for a vector $\boldsymbol{x} = \begin{bmatrix} \boldsymbol{p}_{G,O}^T & r\end{bmatrix}^T$ of 4 unknowns, so the equations can be concatenated and solved for $\hat{\boldsymbol{x}}$ as shown in (\ref{eq:solve_center}), where the pseudo-inverse minimizes the least-squares prediction error between the two sensors. 

\begin{equation} \label{eq:solve_center}
\hat{\boldsymbol{x}} = 
\begin{bmatrix}
\hat{\boldsymbol{p}}_{G,O} \\
\hat{r}
\end{bmatrix} =
\begin{bmatrix}
\boldsymbol{I}_3 & \boldsymbol{n}_{C_f} \\
\boldsymbol{I}_3 & \boldsymbol{n}_{C_a} \\
\end{bmatrix}^{\dagger}
\begin{bmatrix}
\boldsymbol{p}_{G,C_f} \\
\boldsymbol{p}_{G,C_a}\\
\end{bmatrix}
\end{equation}

Using the estimated object radius from (\ref{eq:solve_center}), the desired gripper angle $q_{g,s}^*$ for antipodal contact can be calculated. The function for the distance between the fingertips $d=2r=f(q_g)$ is difficult to invert, but its inverse can be approximated by

\begin{equation} 
q_{g,s}^* \approx a(\hat{r}+r_{sensor}+\epsilon_r) + b
\label{eq:r_inverse}
\end{equation}

\noindent where $a=18.76$rad/m and $b=-0.6129$rad, which has an $R^2$ value of $0.9997$ across the range of motion of the gripper. A small value $\epsilon_r \geq 0 $ is added to the radius estimate to add clearance between the fingertip and the object during the re-grasping trajectory. 

Using the calculated angle $q_{g,s}^*$, the new desired antipodal contact frames relative to the gripper base frame are calculated as $\boldsymbol{T}_{G,C_f^*}$ and $\boldsymbol{T}_{G,C_a^*} = f(q_{g,s}^*)$. The desired object center relative to the gripper base, which is directly between the two fingertips, is given by

\begin{equation}
\boldsymbol{p}_{G,O^*} = \boldsymbol{p}_{G,C_f^*} + \frac{\boldsymbol{p}_{G,C_a^*}-\boldsymbol{p}_{G,C_f^*}}{2}\\
\label{eq:finger_update}
\end{equation}

\noindent where $\boldsymbol{p}_{G,C_f^*}$ and $\boldsymbol{p}_{G,C_a^*}$ are the translations associated with frames $\boldsymbol{T}_{G,C_f^*}$ and $\boldsymbol{T}_{G,C_a^*}$, respectively. Note that $\boldsymbol{p}_{G,O^*}$ is equivalent to $\boldsymbol{p}_{G^*,O}$, the object center location relative to the desired gripper base location, since the object location is assumed to be constant. These calculations are visualized in Fig. \ref{fig:regrasp_diagrams}(b). Transformations can be constructed for both $\boldsymbol{p}_{G,O}$ and $\boldsymbol{p}_{G,O^*}$ according to

\begin{equation} 
\begin{split}
\boldsymbol{T}_{G,O} &=
\begin{bmatrix}
\boldsymbol{I}_3 & \boldsymbol{p}_{G,O} \\
\boldsymbol{0}^T & 1 \\
\end{bmatrix}  \\
\boldsymbol{T}_{G,O^*} &=
\begin{bmatrix}
\boldsymbol{I}_3 & \boldsymbol{p}_{G,O^*} \\
\boldsymbol{0}^T & 1 \\
\end{bmatrix}.  
\end{split}
\label{eq:Tgcents}
\end{equation}

To ensure that the new finger contacts are antipodal, the arm may need to rotate the gripper around the object during the re-grasping trajectory, as illustrated in Fig. \ref{fig:regrasp_diagrams}(c). Two correction angles, $\theta'$ and $\phi'$, are calculated and used to construct a rotation that is applied between the estimated contact center frame and the desired contact center frame, $\boldsymbol{T}_{O,O^*}$.

\begin{equation} \label{eq:correction_angles}
\theta' = \frac{\theta_f + \theta_a}{2}, \quad \phi' = \frac{\phi_f - \phi_a}{2}
\end{equation}

\begin{equation}
\boldsymbol{T}_{O,O^*} = \boldsymbol{T}_{G,C_f^*}^{-1}\boldsymbol{T}_{G,C_f}
\begin{bmatrix}
\boldsymbol{R}_y(\phi')\boldsymbol{R}_x(\theta') & \boldsymbol{0} \\
\boldsymbol{0}^T & 1 \\
\end{bmatrix} 
\label{eq:gripper_rot}
\end{equation}

Finally, the transformation for the desired gripper base pose can be calculated as

\begin{equation} 
\boldsymbol{T}_{W,G^*} = \boldsymbol{T}_{W,G}\boldsymbol{T}_{G,O}\boldsymbol{T}_{O,O^*}\boldsymbol{T}_{G,O^*}^{-1}
\label{eq:gripper_desR}
\end{equation}

\noindent where $\boldsymbol{T}_{W,G}$ is the current pose of the gripper base in the world frame. Using $\boldsymbol{T}_{W,G^*}$, the inverse kinematics for the arm can be calculated to obtain new joint angles $q_{f,i}^*$ that will achieve the desired gripper pose. 

Due to large angle changes in inverse kinematics solutions depending on $\boldsymbol{T}_{O,O^*}$, it was not feasible to directly implement the returned solutions. To approximate the solution, the rotation was decomposed into Euler angles, with the first rotation, $\alpha$, being about the wrist-roll joint axis. This specific case is shown in Fig. \ref{fig:regrasp_diagrams}(c) and (d). Equation (\ref{eq:gripper_desR}) is modified so that $\boldsymbol{T}_{W,G^*}$ is then given by

\begin{equation} 
\boldsymbol{T}_{W,G^*} = \boldsymbol{T}_{W,G}\boldsymbol{T}_{G,O}\boldsymbol{T}_{G,O^*   }^{-1}
\label{eq:gripper_des}
\end{equation}

\noindent with an updated inverse kinematics solution. The rotation $\alpha$ is directly added to the $q_{f,6}^*$ value returned by the new inverse kinematics solution.

For each of the follower arm joints, the re-grasping trajectory is constructed as a linear interpolation between the current joint angle $q_{f,i}$ and the final joint angle as determined by the inverse kinematics $q_{f,i}^*$ over the total trajectory time, $T_f$. The gripper trajectory is a linear interpolation between the current gripper position and angle calculated in (\ref{eq:r_inverse}). During the trajectory, the leader arm tracks the follower arm motion using the PD control law in (\ref{eq:teleop_coupling}). The total trajectory time is set to 150\si{ms}, but the hardware allows for even faster trajectories. An overhead view of an example re-grasp is shown in Fig. \ref{fig:arm_with_timelapse}.

After the re-grasping trajectory has been executed, the grasp detection conditions are evaluated again. If the re-grasp trajectory was successful, the anti-slip reflex is activated. If the trajectory was unsuccessful, possibly due to noisy object estimates or unintentional contact with the object, the re-grasping reflex will be re-activated and a new trajectory will be planned and executed.

\begin{table}[t]
\caption{Algorithm Parameters}
\label{tab:alg_params}
\centering
\begin{tabular}{ c  c  c } 
 \hline
Purpose & Symbol & Value \\
 \hline\hline
Polar angle threshold & $\gamma_{\psi}$ & 0.3rad \\ 
 \hline
Normal force threshold & $\gamma_{n}$ & 0.3N \\ 
 \hline
Estimated coefficient of friction & $\hat{\mu}$ & 0.5 \\ 
 \hline
Friction cone threshold & $\gamma_{c}$ & 1.6 \\ 
 \hline
Radius increase for trajectory & $\epsilon_r$ & 0.01m \\ 
 \hline
Re-grasp trajectory duration & $T_{f}$ & 0.015s \\ 
 \hline
 \\
\end{tabular}
\vspace{-6mm}
\end{table}

\begin{figure}[t]
\centering
\includegraphics[width=0.95\linewidth]{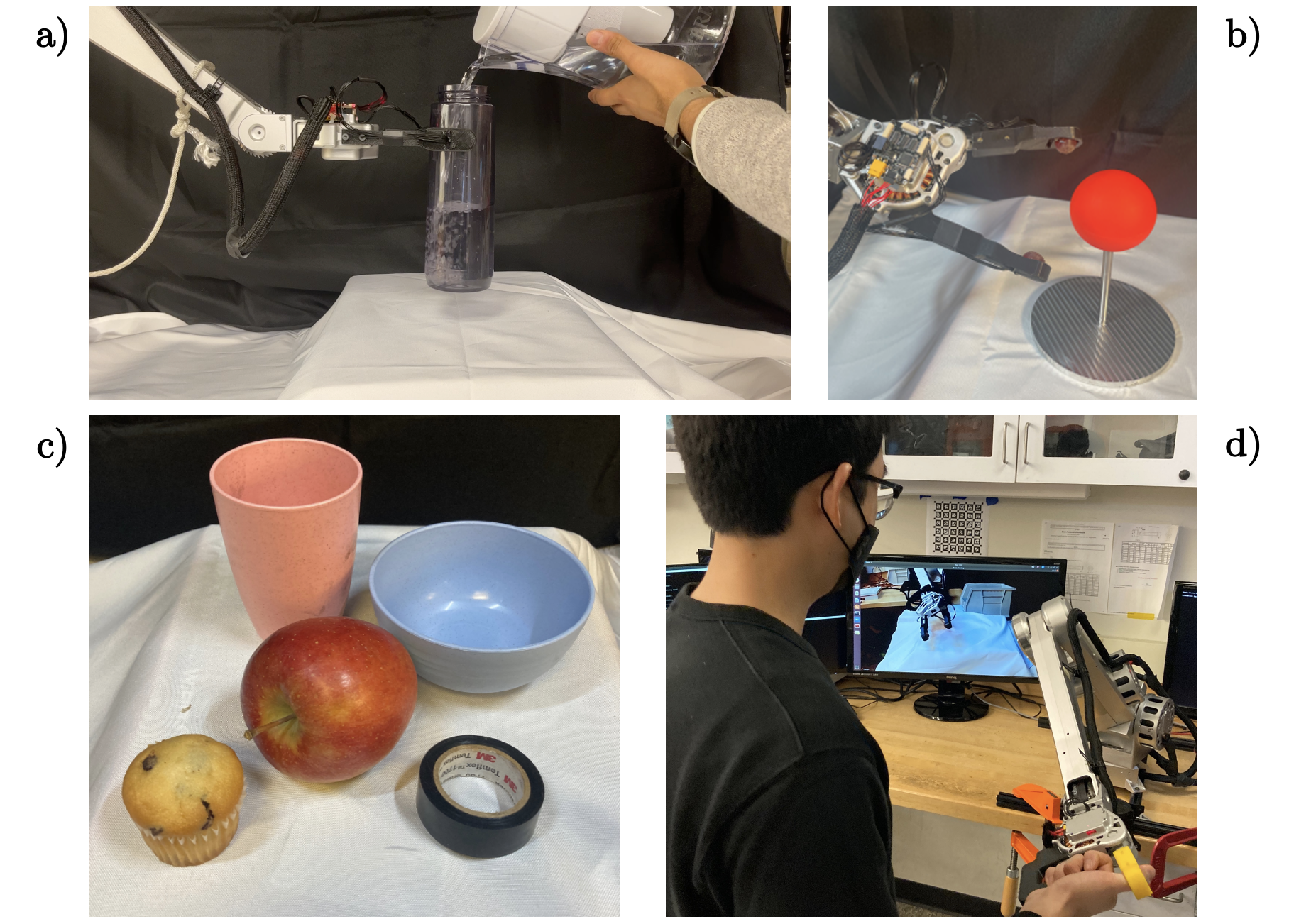}
\caption{\textbf{Experimental setups.} (a) Filling a water bottle to test slip control reflex. (b) Fixed sphere to test re-grasping reflex. (c) Various objects to evaluate teleoperation performance on pick-and-place tasks. (d) Input setup for teleoperation tasks. The user holds the leader gripper and observes the follower workspace through a USB camera.}
\label{fig:experimental_setup}
\vspace{-2mm}
\end{figure}

\section{Experimental Studies} \label{sec:results}

Two separate experiments to validate the anti-slip and re-grasping reflexes are discussed in Sections \ref{sec:results:slip} and \ref{sec:results:regrasp}, respectively. To test the combined performance of the reflexes, several hours of teleoperation data were also collected for different pick-and-place style tasks. This dataset is presented and analyzed in Section \ref{sec:results:teleop}. The algorithm parameters used in our experiments are shown in Table \ref{tab:alg_params}.

\subsection{Anti-Slip Reflex} \label{sec:results:slip}

\begin{figure}[t]
\centering
\includegraphics[width=0.9\linewidth]{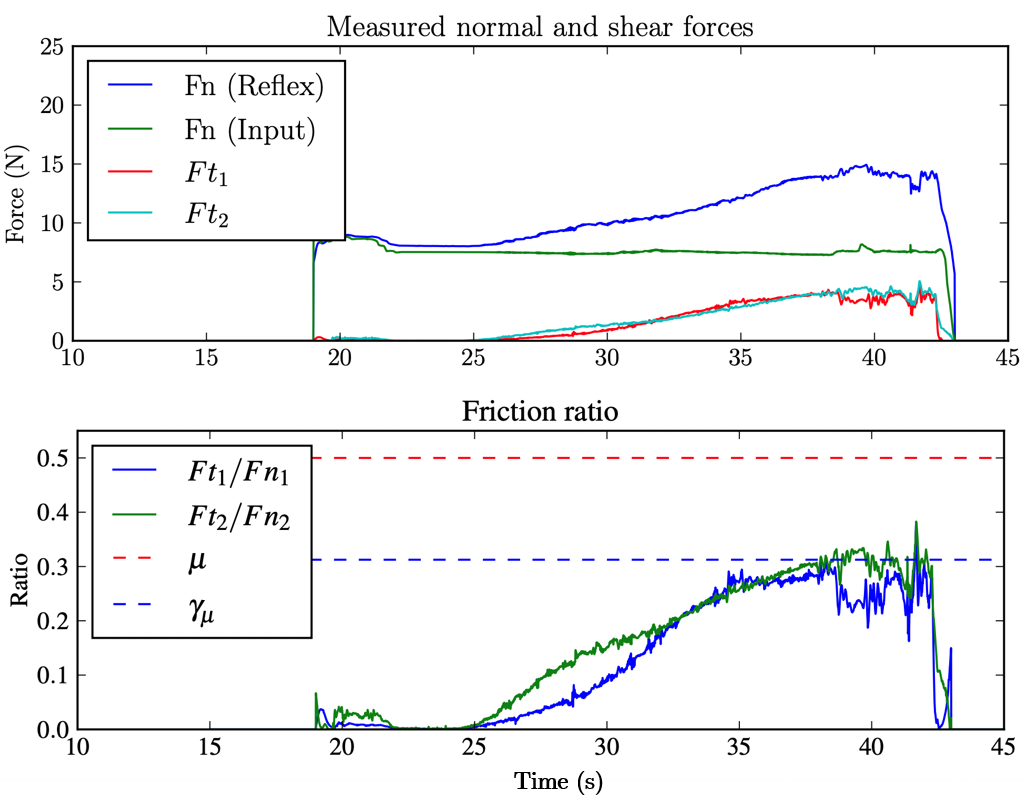}
\caption{\textbf{Slip control while filling the water bottle.} As the water bottle is filled, the sensed shear force increases. The reflex controller increases the normal force to prevent slip by increasing the gripper torque. The controller successfully keeps the resulting forces away from the friction threshold and thus the edge of the friction cone.}
\label{fig:slip_results}
\vspace{-4mm}
\end{figure}

To evaluate the response of the anti-slip reflex, a teleoperator lightly holds a water bottle as it is filled up, as shown in Fig. \ref{fig:experimental_setup}(a). As the bottle is filled, the user applies a constant force while the reflex controller supplies the necessary additional normal force to prevent slip. The upper plot in Fig. \ref{fig:slip_results} shows the evolution of the measured shear forces, the applied normal force, and the user input force. By the end of the trial, the anti-slip reflex is applying nearly half of the normal force. The lower plot of Fig. \ref{fig:slip_results} shows how the slip controller keeps the resulting force ratio below the threshold value $\gamma_\mu$, thus keeping the total forces on the bottle within their respective friction cones.

\subsection{Re-Grasping Reflex} \label{sec:results:regrasp}

\begin{figure}[t]
\centering
\includegraphics[width=0.95\linewidth]{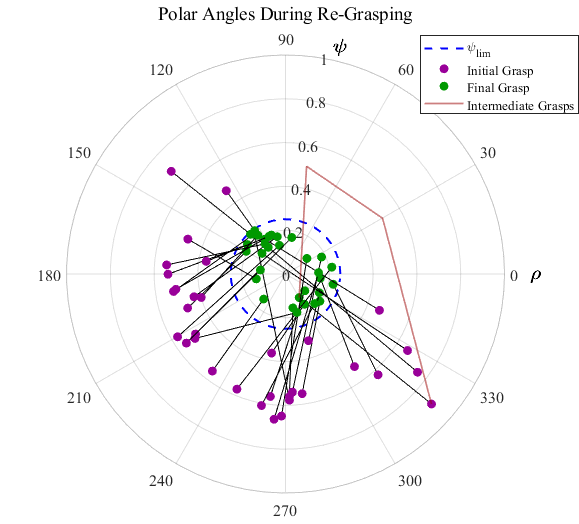}
\caption{\textbf{Initial and final values of $\psi$ during re-grasping.} Plot of polar angle $\psi$ vs. azimuthal angle $\rho$ of contacts on the sensor surface for the re-grasping reflex test. Dark grey lines connect the corresponding initial and final grasp locations. A single sequence of re-grasping attempts is shown in red. From varied initial grasp locations on the sensor surface, the re-grasping reflex successfully achieves antipodal contact angles in the final grasp. Due to the consistent approach direction between the gripper and the objects, the initial grasp locations on the sensor surface are not uniformly distributed. The polar angles of the initial grasps have a mean of 0.39rad and a standard deviation of 0.10rad. The polar angles of the final grasps have a mean of 0.18rad and a standard deviation of 0.04rad. 
}
\label{fig:regrasp_psi}
\end{figure}

To evaluate the re-grasping reflex, a solid ball is fixed at a constant position in the arm workspace, as shown in Fig. \ref{fig:experimental_setup}(b). The user approaches the ball using different gripper orientations, initiating grasp attempts from different regions of the sensor surfaces. Fig. \ref{fig:regrasp_psi} shows the initial and final grasp positions on the sensor surface, parameterized by the polar angle $\psi$ from (\ref{eq:psi_calc}) and the azimuthal angle $\rho$. The azimuthal angle is given by

\begin{equation}
\rho = \text{atan2}(n_y,n_x),
\label{eq:rho_calc}
\end{equation}

\noindent where $\textbf{n} = [n_x,n_y,n_z]^T$ is the contact frame's normal vector. Due to sensor noise or collisions between the ball and the sensor fingers, some grasps required multiple re-grasp attempts to establish antipodal contact. Despite this, a successful final grasp was obtained for every initial grasp position. An example series of re-grasp attempts is shown as a red line in the figure. 
Across all of the grasps, the median time from initial contact detection to the final grasp was 0.55s, and the median number of attempted grasps was 2.

\begin{table}[t]
\caption{Regrasp performance across user trials}
\label{tab:regrasp_performance}
\centering
\begin{tabular}{c c c c c c} 
 \hline
Controller & Initial $\psi$ & Initial $\psi$ & Final $\psi$ & Final $\psi$ \\ 
& Mean & SD & Mean & SD \\
& (rad) & (rad) & (rad) & (rad) \\
 \hline\hline
 Teleoperation & 0.40 & 0.11 & -- & -- \\ 
 \hline
 Reflexive grasping & 0.47 & 0.22 & 0.16 & 0.04 \\ 
 \hline
\end{tabular}
\vspace{-4mm}
\end{table}

\begin{figure*}[t]
\centering
\includegraphics[width=0.95\linewidth]{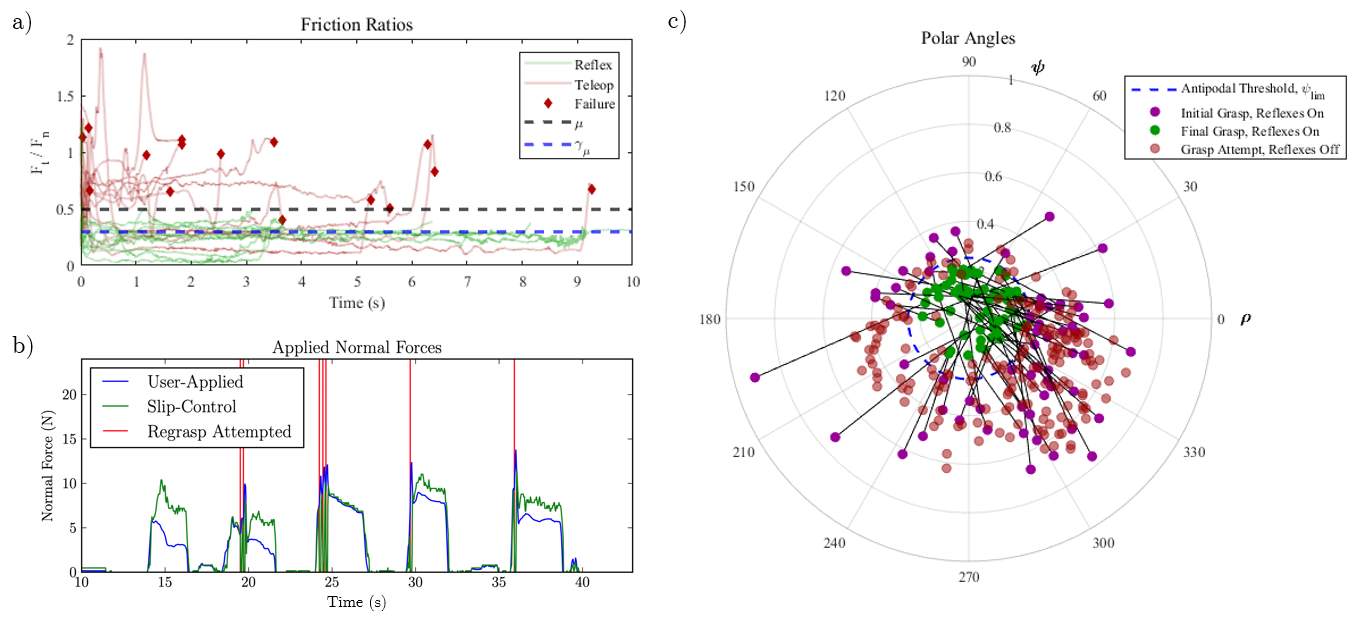}
 \caption{ \textbf{Visualizing reflex performance}. (a) Example plot of friction ratios over a subset of user trials, with and without the grasping reflexes enabled. Without the reflexes, the object is much more likely to slip out of attempted grasps. (b) Example of sequential grasping attempts with the grasping reflexes enabled. Despite imperfect user inputs, different combinations of reflexes are used to ensure the grasps are successful. (c)  Polar plot showing grasp locations on sensor surface over all user trials, with and without the grasping reflexes enabled. Very few of the initial grasps during teleoperation were within the antipodal threshold, while all of the reflexive grasps ended up within the threshold.  
 }
\label{fig:user_data}
\vspace{-4mm}
\end{figure*}

\subsection{Teleoperation for Pick-and-Place Tasks} \label{sec:results:teleop}

Teleoperated pick-and-place trials were conducted with and without the reflexes enabled, using the objects shown in Fig. \ref{fig:experimental_setup}(c). The objects were arranged on the table in groups of four at a time, with an individual trial corresponding to the user picking a single object. An image of the leader setup is shown in Fig. \ref{fig:experimental_setup}(d), where the user operates the leader device while watching the follower workspace through a USB camera with no added latency. Trials were conducted with five novice users, who each had less than ten minutes of practice with the system, and a single expert user, who had over two hours of prior experience with the system. 
The friction ratio for a set of attempted grasps is shown in Fig. \ref{fig:user_data}(a) and a series of sequential grasping attempts with the reflexes activated is shown in Fig. \ref{fig:user_data}(b). Across all of the attempted grasps, the polar angles are shown in Fig. \ref{fig:user_data}(c). 

During teleoperation without the grasping reflexes, failed grasp attempts can be seen as spikes in the friction ratio plot which occur as the gripper loses contact with the objects. In contrast, the ratios for the reflexive grasping attempts all stay below the friction cone boundary. This indicates that the user had control over the object until the moment they chose to release the leader gripper and disable the slip control reflex.

The improvement in grasp quality with enabled reflexes can also be seen in the plot of the polar contact angles. Similar to Fig. \ref{fig:regrasp_psi}, the pre- and post-re-grasping angles are plotted. The initial contact angles for each grasping attempt without the reflexes are also plotted. Statistics for the re-grasp performance can be seen in Table \ref{tab:regrasp_performance}. While the standard deviations of the distributions for the initial contact angles are similar, only a few of the non-reflexive grasping attempts have polar angles within the threshold $\psi_{lim}$. All of the final grasps for the reflex-enabled trials were within the threshold, and the variance of the final grasp angles is much lower.

A summary of the pick-and-place tasks is presented in Table \ref{tab:reflex_compare}.  While an identical number of pick-and-place trials were performed with the reflexes disabled and enabled, the reflexes greatly reduced the number of user grasp attempts required to complete the task. The success rate, defined as the ratio of successful to total grasp attempts within each trial, is nearly 100\% when the reflexes are enabled, and the average time per successful pick-and-place trial is reduced by 26\%. For context, our pick-and-place trial times are compared with other teleoperation studies in Table \ref{tab:time_comparison}. The average pick-and-place time for novice users of our system was roughly 40\% lower than average trial times reported in a study where the user received high-fidelity haptic feedback \cite{khurshid2016effects} and over 85\% faster than the average trial times reported in studies where the user only received visual feedback \cite{meeker2018intuitive,handa2020dexpilot}. Although these comparisons are made across different experimental setups and conditions, there is a stark difference in trial times for picking and placing individual objects.

\begin{table}[t]
\caption{Teleoperation Performance with Added Reflexes}
\label{tab:reflex_compare}
\centering
\begin{tabular}{c c c c c c} 
 \hline
Controller & Trials & Grasp & Success & Avg. $t$ \\ 
 & & Attempts  & Rate & (s) \\
 \hline\hline
 Expert teleoperation & 16 & 31 & 52\% & 2.78 \\ 
 \hline
 Expert w/ reflexes & 16 & 16 & 100\% & 1.74 \\ 
 \hline
 Novice teleoperation & 84 & 178 & 47\% & 4.99 \\ 
 \hline
 Novice w/ reflexes & 84 & 86 & 98\% & 3.70 \\ 
 \hline
\end{tabular}
\end{table}

\begin{table}[t]
\caption{Comparing Teleoperated Pick-and-Place Times}
\label{tab:time_comparison}
\centering
\begin{tabular}{c c c} 
 \hline
 & Mean & \% Improvement \\ 
  Source & Completion & (Proposed (naive) \\
 & Time (s) & from Reference) \\
 \hline\hline
Proposed reflexes (expert) & 1.74 & - \\
\hline
Proposed reflexes (naive) & 3.70 & - \\ 
 \hline
Reference 1 (Khurshid \cite{khurshid2016effects}) & $\sim6.5$ & 43\% \\ 
 \hline
Reference 2 (Meeker \cite{meeker2018intuitive}) & 27.52 & 87\% \\ 
 \hline
Reference 3 (Handa \cite{handa2020dexpilot}) & $\sim25$ & 85\% \\ 
 \hline
\end{tabular}
\vspace{-4mm}
\end{table}

\section{Discussion} \label{sec:disc}

The proprioceptive manipulation hardware and bimodal fingertip sensors used in this work are key to achieving high-speed teleoperation performance, but the autonomous reflexes could still be deployed on a traditional system (as long as the contact location and force information is available). 
In this case, the manipulation platform will be slower, and while the reflexes will still be able to improve the robustness of grasping they will not be able to improve the speed.
Even in this work, the force sensors and gripper design impose some limitations on the reflexes and the teleoperation performance. The sensors have a sample rate of just 200Hz and a minimum force threshold of around 0.5N, and the gripper only has one degree of freedom. As sensing capabilities are improved, and the gripper becomes more capable, the reflexes shown in this work can be even more responsive. Furthermore, with increased sensing capabilities more complex and diverse libraries of these reflexive controllers can be created.

Additionally, while this work uses a human-in-the-loop to provide the coarse manipulation instructions, such as locating and initiating contact with the object, this does not need to be the case. Both of the reflexes presented here could be easily integrated with a traditional computer-vision-based manipulation system, potentially circumventing the more computationally intense grasp-selection algorithms. Investigating how to combine our reflex controllers with modern manipulation planners is planned for future work.

Both of our proposed reflexes are built around the assumption that grasped objects are circular or close to circular. This assumption holds for many common objects and is a reasonable starting point, but it will need to be relaxed in future work. Most initial grasps on rectangular objects will have antipodal finger contacts, but triangular objects will be particularly adversarial. Characterizing the performance of these reflexes on non-circular objects and designing additional reflexes are important avenues for future work.

\section{Conclusion} \label{sec:conc}
We have presented a proprioceptive teleoperation platform consisting of two six degree-of-freedom arms. The characteristics of the arms and the equipped bimodal force sensors are exploited to design anti-slip and re-grasping reflexes, which improve the speed and accuracy of the teleoperation without requiring any changes to the user inputs.

\section*{ACKNOWLEDGMENTS}

The authors thank Menglong Guo and Aditya Mehrotra for their hardware support and Steve Heim and Hongmin Kim for their insightful discussions.
This work was supported by the Advanced Robotics Lab of LG Electronics Co., Ltd.

\addtolength{\textheight}{-7cm}

\bibliographystyle{IEEEtran} %
\bibliography{library}

\end{document}